\documentclass{article}

\usepackage{microtype}
\usepackage{graphicx}
\usepackage{subfigure}
\usepackage{booktabs} 

\usepackage{hyperref}

\usepackage[accepted]{icml2024}

\usepackage{amsmath}
\usepackage{amssymb}
\usepackage{mathtools}
\usepackage{amsthm}
\usepackage{csquotes}
\usepackage{environ}
\usepackage[capitalize,noabbrev]{cleveref}

\theoremstyle{plain}

\theoremstyle{definition}

\theoremstyle{remark}

\usepackage[textsize=tiny]{todonotes}

\newcommand{\llm}[1]{\textcolor{brown}{#1}}
\newcommand{\symbolic}[1]{\textcolor{teal}{#1}}

\NewEnviron{prompt}{
\begin{displayquote}
\texttt{\BODY}
\end{displayquote}
}

\icmltitlerunning{Modeling Referential Expression Generation with Scaffolded LLMs}

\begin{document}

\twocolumn[
\icmltitle{Cognitive Modeling with Scaffolded LLMs: A Case Study of Referential Expression Generation}



\icmlsetsymbol{equal}{*}

\begin{icmlauthorlist}
\icmlauthor{Polina Tsvilodub}{tue}
\icmlauthor{Michael Franke}{tue}
\icmlauthor{Fausto Carcassi}{amst}
\end{icmlauthorlist}

\icmlaffiliation{tue}{Department of Linguistics, University of T\"ubingen, T\"ubingen, Germany}
\icmlaffiliation{amst}{ILLC, University of Amsterdam, Amsterdam, Netherlands}

\icmlcorrespondingauthor{Polina Tsvilodub}{polina.tsvilodub@uni-tuebingen.de}

\icmlkeywords{Machine Learning, ICML}

\vskip 0.3in
]



\printAffiliationsAndNotice{} 

\begin{abstract}
To what extent can LLMs be used as part of a cognitive model of language generation?
In this paper, we approach this question by exploring a neuro-symbolic implementation of an algorithmic cognitive model of referential expression generation by \citet{dale1995computational}. The symbolic task analysis implements the generation as an iterative procedure that scaffolds symbolic and \texttt{gpt-3.5-turbo}-based modules. We compare this implementation to an ablated model and a one-shot LLM-only baseline on the A3DS dataset \cite{tsvilodub-franke-2023-evaluating}. We find that our hybrid approach is cognitively plausible and performs well in complex contexts, while allowing for more open-ended modeling of language generation in a larger domain.
\end{abstract}

\section{Introduction}
Large language models (LLMs) have shown impressive performance on different benchmarks on a variety of tasks \citep[e.g.,][]{NEURIPS2020_1457c0d6,bommasani2021opportunities, chowdhery2022palm, touvron2023llama}. 
Recently, LLMs have been embedded within larger systems, also called LLM \textit{agents}, which can for instance retrieve relevant information \cite{lewis2020retrieval, liu-etal-2022-generated}, make use of additional computational components for math problem solving \cite{he2023solving}, more complex reasoning tasks \cite{creswell2022selection, he2023solving, paranjape2023art, poesia2023certified}, programming tasks \cite{gao2022pal}, or generating better texts through additional computational steps \cite{piriyakulkij2023asking}.
Other recent work has taken a more cognitively inspired perspective using LLMs in hybrid, neuro-symbolic models for extending explanatory cognitive models \citep[e.g.,][]{lew2020leveraging} or as part of cognitive architectures \citep[e.g.,][]{sumers2023cognitive,wong2023word}.
Here, rather than focusing on LLM performance we take the perspective of cognitive scientists, focusing on task analysis with the goal of building computational models of cognitive processes. In this perspective, we pose the question: To what extent can LLMs be used as components in implementation of general algorithmic models of cognitive processes?

We focus on linguistic cognition and pick out a minimal non-trivial case study of \textit{referential expression} generation in a \textit{contrastive reference game task}. 
This task requires the production of a description to identify a \textit{target} referent among a set of possible alternative \textit{distractor} referents. 
Several prominent approaches in computational linguistics and cognitive science treat  reference games, essentially, as a mapping problem.
Given a set of distractors, the target, and a fixed list of expressions (with a known semantics), the task is construed as selecting the best, or a good enough, expression to describe the target \cite{KramerDeemter2012:Computational-G,frank2012predicting}. 
While these approaches have successfully captured that cooperative speakers exploit pragmatic reasoning rooted in Gricean Maxims to convey the intended content while keeping utterances brief \cite{grice1975logic}, they require a pre-specified list of, or a construction procedure for, the descriptions to select from, which essentially reduces the task to the problem of mapping utterances onto states. This ``closed world'' problem is shared by many instances of whole classes of approaches, including many types of Bayesian models or model-based reinforcement learning.

Several neuro-symbolic models for reference games have been proposed to overcome this limitation. 
Some use pragmatic agent models to learn suitable semantic meaning representations \citep{MonroePotts2015:Learning-in-the,OhmerFranke2021:Mutual-Exclusive-CogSci}. 
Others train neural networks to perform well on the problem of in-context discrimination \citep{MaoHuang2016:Generation-and-,HendricksAkata2016:Generating-Visu}.
More cognitively inspired models build on extant probabilistic models \citep{frank2012predicting}, adding modules for neural language generation 
\citep{AndreasKlein2016:Reasoning-about,VedantamBengio2017:Context-Aware-C,Cohn-GordonGoodman2018:Pragmatically-I,NieCohn-Gordon2020:Pragmatic-Issue,ZarriessSchlangen2019:Know-What-You-D}.
However, all these require task-specific training or appropriately pre-trained base models. Moreover, even if inspired by extant cognitive models, they often do not aspire to remain faithful to the original in favor of achieving higher task accuracy.

In this work, we explore how successfully modern LLMs can be used to overcome this ``closed-world'' problem for known algorithmic solutions to the reference game task. 
Specifically, we consider a simple algorithmic idea from \citet{dale1995computational} and extend it with LLM modules to make it more widely applicable without requiring hand-specified, closed sets of utterance alternatives. We critically assess this neuro-symbolic model against an ablation and a baseline. 

\begin{figure*}[ht]
\vskip 0.2in
\centering
    \centerline{\includegraphics[width=0.8\textwidth]{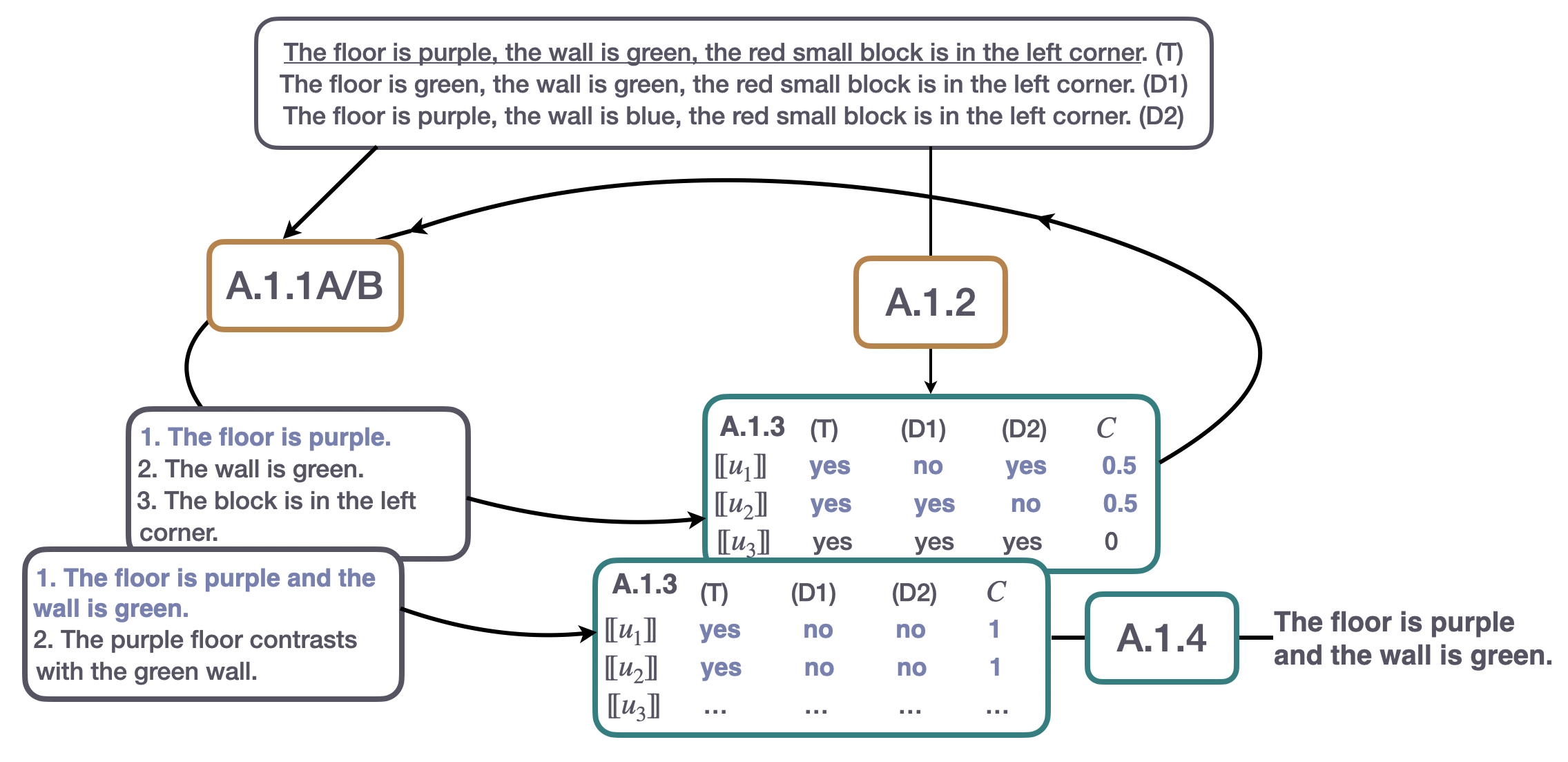}}
    \caption{The figure shows two iterations of the model, ending with the production of a contrastive utterance. T, D1, and D2 denote the target and the two distractor states respectively. $C$ indicates the contrastivity values. Only the target is passed to the utterance proposer. Boxes with a \llm{brown} border indicate LLM-based modules, components with \symbolic{green} symbolic modules. The labels of the modules indicate section numbers in the appendix containing the full details.
    \label{fig:ref-games-example}}
\end{figure*}

\section{Model \& Experiments}

\subsection{Iterative Model}

We build on the Incremental Algorithm (IA) by \citet{dale1995computational}, in which a referential expression is constructed by iterating through a pre-specified list of attributes and adding each attribute that applies to the target and rules out at least one distractor.
While the IA is relatively simple and more performant solutions exist, ordering the attributes according to human preferences is psycholinguistically motivated. Moreover, the general idea of incremental search algorithms has a long history \cite{NewellSimon1972a}, demonstrable advantages also for LLM-based architectures \cite{yao2023tree}, and is supported in the particular domain of language generation \citep{Ferreira2019:A-Mechanistic-F}.

The IA of \citet{dale1995computational} requires \textit{a priori} specification of (i) the order in which the attributes are considered, (ii) all possible states and their attribute values, (iii) a procedure for utterance construction, as well as (iv) semantics of utterances applied to any state, all of which may be domain-specific. 
To address these limitations, we propose a variation of the IA which combines the IA's symbolic computations with LLM generation.

Our \textit{iterative model} (IM) is illustrated in Figure~\ref{fig:ref-games-example} (IM; see Algorithm~\ref{alg:iterative-model} in the appendix for full details). 
The model takes as input a context which consists of a target state $s^*$ and one or more distractors $D$.
It outputs a contrastive description for $s^*$.
It uses an LLM for (i) proposing descriptions (module: \textit{UtterancesProposer}) and (ii) evaluating whether a proposed description is semantically compatible with any state (target or distractor, module: \textit{SemanticEvaluator}).\footnote{
For all LLM components and the baseline, GPT-3.5 (\texttt{gpt-3.5-turbo}, checkpoints of summer 2023) with temperature $\tau=0.1$ was used.
All LLM prompts and details on functionality and evaluation of the single modules can be found in the Appendix~\ref{app:section}.}
Symbolic components subsequently (i) check for contrastivity of each proposed description, and (ii) potentially iterate the procedure if no fully contrastive description has been found.

Contrastivity is defined as the proportion of distractors for which the utterance is false. For instance, if there are two or four distractors and the utterance is false of one, the contrastivity is 0.5 or 0.75, respectively (see Fig.~\ref{fig:ref-games-example}).
At each iteration, if none of the utterances is fully contrastive (i.e., uniquely identifies the target), the symbolic component selects the most informative utterances available (i.e., the most contrastive ones), and passes them to the UtterancesProposer again, which is prompted to add some detail about the target to the utterance based on the previous utterance, and starts a new iteration.\footnote{Note that distractors are not taken into account when extending the description of the target. Instead, we consider only the utterances produced so far along with the target. This potentially makes the module more task-agnostic and reusable for future applications. Note that the UtterancesProposer in Alg.~\ref{alg:iterative-model} uses a different initial prompt for the first iteration; see App.~\ref{app:im-utt-proposer}.}
This is repeated until an utterance is found that solves the task or the maximal number of iterations (in our simulations, five) is achieved. The most contrastive utterance is greedily selected and returned.

An example of a pass through the IM can be found in Figure~\ref{fig:ref-games-example} (to be read counter-clockwise, starting at the top and ending at the bottom right). Example inputs for one target and two distractors are shown in the top box. The first iteration starts with the target description (marked with ``T'') being passed to the UtterancesProposer (A.1.1A/B); example outputs are shown in the next box (e.g., ``The floor is purple''). The sampled utterances are then passed to the SemanticEvaluator (A.1.3, shown as columns) evaluated for all states (shown as rows). Results of evaluating each utterance for each state are shown in the cells, with column $C$ showing the contrastivity values. The more contrastive utterances (i.e., those setting the target apart from one distractor) are not fully contrastive, so they are passed to the UtterancesProposer (A.1.1A/B) again, and the model is prompted to generate more detailed samples (e.g., ``The floor is purple and the wall is green''). The second iteration through the model is computed and since a fully contrastive extended utterance is found (i.e., with $C=1$, two utterances in this example), it is returned as the final expression (A.1.4). 

The IM implements a search over a tree of possible referential utterances \citep[cf.,][]{yao2023tree}, which has been a long-standing approach in AI \citep[][]{van2023reclaiming}.
Since each iteration adds a single detail about the target, in an order proposed by the LLM itself rather than a manually-specified order like in the original IA, the tree depth roughly corresponds to the number of details included in the generated utterance, while the width corresponds to the number of sampled utterance proposals on each iteration. 
The IM implements a partial search, by passing only the currently most contrastive utterances to the subsequent iteration.

The modules used for such computational models might generally fall into \textit{functionally different types} \citep[cf.,][]{sumers2023cognitive}. \emph{Evaluators} provide context-dependent assessment of alternatives, \emph{proposers} supply these possible alternatives or contingencies (e.g., plausible utterances for a given context), and symbolic or simply rule-based (as in this case) modules which can generally be called \emph{selectors} combine and process information supplied by the other two types of modules. Given their context-dependence, the first two types will often be neural.

\subsection{Ablated Model}

To test the impact of iteration in the IM, we compare it to an ablated \textit{single-pass} model (SP; Algorithm~\ref{alg:ablated-model}), which samples utterances, evaluates them and selects the best alternative in a single pass, without iteration.
%
We expect the performance of the SP model to scale with the number of sampled utterances. Maintaining a very large number of alternatives may be cognitively implausible due to resource constraints, lending credence to more iterative approaches \citep{Ferreira2019:A-Mechanistic-F}.

\subsection{Baseline}
We compare the results of the two neuro-symbolic models (IM and SP) with a baseline model. 
The baseline consists of a single call to the LLM asking for an utterance that solves the task.
We use a one-shot chain-of-thought prompt, shown in Appendix~\ref{app:baseline}, \citep[][among others]{wei2022chain}.


\subsection{Simulations}
We test the models for reference games based on a derivative of the 3Dshapes dataset \cite{burgess20183d}, called A3DS \cite{tsvilodub-franke-2023-evaluating}.
A3DS contains textual descriptions of scenes consisting of a 3D geometric object in an otherwise empty room. 
All scenes in the used subset are unique and consist of a combination of the following attributes: shape of the object (four values); size of the object (three values); color of the wall, the floor, the object itself (independent of each other, seven possible values); object position relative to the background (three values). 
The scene descriptions used as input to the models were of the form ``The floor is \{floor color\}, the wall is \{wall color\}, the \{object color\} \{size\} \{object\} is in the \{position\}.'' Example inputs are shown in Figure~\ref{fig:ref-games-example} (top box).
This state-space is highly structured and would, in principle, allow the specification of a large description set with a grammar and compositional semantic rules.
We use this data set not because it provides an insurmountable obstacle to the original IA, but rather because it simplifies the evaluation of the models' outputs, allowing us to automatically verify whether a generated utterance mentions contrastive features and calculate the overall contrastivity for a given context, without human annotation.

To compare the performance of  the IM, SP and baseline models, we construct reference games by first sampling a target state, and then adding one, four or eight distinct distractors. Each distractor differs from the target by maximally two features. In the IM, the UtterancesProposer sampled either four or eight utterances. For the SP, the UtterancesProposer always sampled ten utterances. 
We tested all models on 100 reference games for each of the parameter configurations.
\begin{figure}[ht]
\vskip 0.2in
\centering
\centerline{\includegraphics[width=\columnwidth]{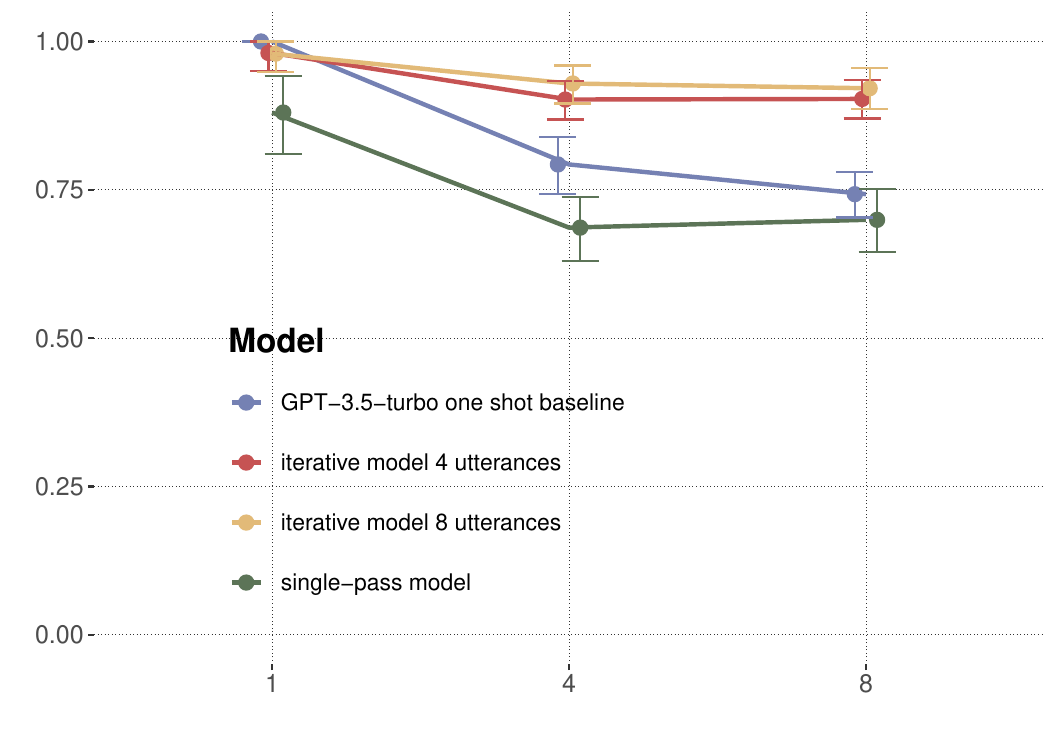}}
    \caption{Distribution over contrastivity values (y-axis) by number of distractors (x-axis) and number of utterances proposed (color). Error bars show bootstrapped 95\%-CIs.
    \label{fig:ref-games-summary}}
\vskip -0.2in
\end{figure}
\begin{figure}[ht]
\vskip 0.2in
\centering
\centerline{\includegraphics[width=\columnwidth]{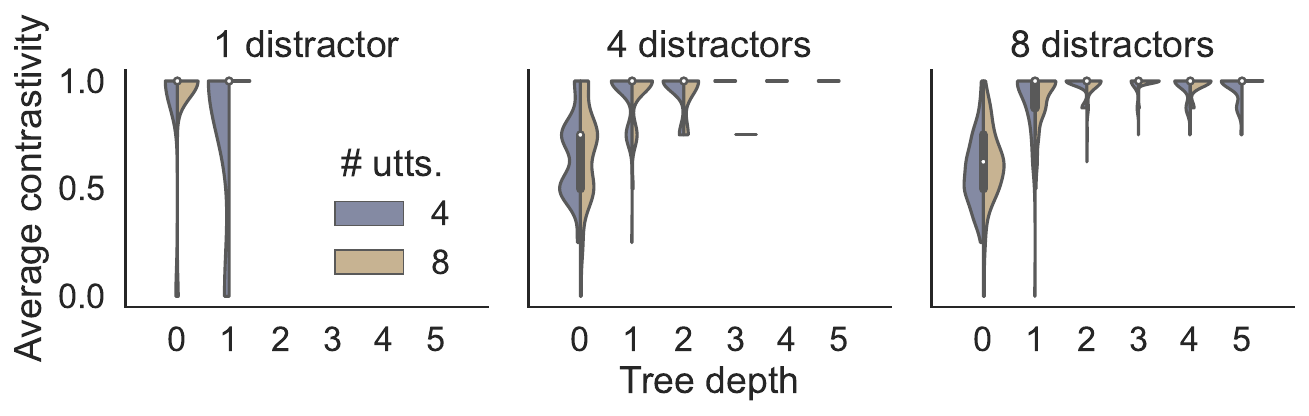}}
    \caption{Development of task success over increasing tree depth in the IM: distribution over contrastivity values (y-axis) over increasing tree depth (extended utterance proposal and evaluation iterations; x-axis), by number of distractors (facets) and tree width (number of proposed utterances; color). Dots indicate means, thick bars indicate quartiles, thinner lines indicate minimal values.
    \label{fig:ref-games-loops}}
\vskip -0.2in
\end{figure}
\subsection{Results} 
Model performance is measured by average contrastivity across references games (Figure~\ref{fig:ref-games-summary}).
The IM generated highly contrastive utterances which successfully set apart the target from the distractors, even with more distractors, outperforming the baseline and the SP.
The average contrastivity of the IM for four and eight distractors was above the baseline (bootstrapped $P=1$), while the difference was not significant for one distractor due to a ceiling effect.
The performance of the ablated SP model decreased as the number of distractors in the context increased. The performance of the ablated SP model was credibly worse than all other models, across the number of distractors (bootstrapped $P=1$).
The LLM baseline outperformed the SP.

The number of iterations required for the IM until fully contrastive utterances were produced increased with the number of distractors and thus with the difficulty of identifying contrastive features (Figure~\ref{fig:ref-games-loops}). In particular, this shows that the IM increased the complexity of the computation and of the generated utterances in a context-dependent way.
Furthermore, Figure~\ref{fig:ref-games-loops} suggests a slight trade-off between the tree width and tree depth required for producing contrastive utterances (x-axis~vs.~color): when more utterances were proposed at each step, it was more likely that at least one of them mentioned contrastive features, so that fewer iterations were required overall. 

\section{Discussion}
In this paper, we focused on contrastive utterance generation in a reference game setting. We implemented an iterative neuro-symbolic model based on the IA \cite{dale1995computational}. The model adapts the generated utterances to the complexity of the task at hand, while producing diverse language that is natural in context, which is enabled by the performance of LLMs. 
We found that the IM outperforms both an ablated single-pass model and an LLM baseline in complex reference games.

The proposed approach has certain advantages over approaches requiring fine-tuning. For instance, \citet{tsvilodub-franke-2023-evaluating} used the A3DS data set to fine-tune a pre-trained image captioner to produce contrastive captions via reinforcement learning.
In a comparable test setting (one distractor), the pragmatic model only achieved an accuracy of 0.54.
Although directly conditioned on visual input rather than ``oracle'' text input, it required costly fine-tuning and did not match the performance shown here.
In contrast, the considered neuro-symbolic approach didn't require fine-tuning, and showed better performance, while being agnostic to the neural model.
Additionally, the IM model offers an algorithmic cognitively motivated (symbolic) task decomposition which requires spelling out assumptions about the modeled process, which scaffolds the black-box LLM modules. 
Such a task analysis allows for easy explorations of a given model via plug-and-play with the modules.
For instance, other search algorithms could be considered, which would also allow to explore more computationally efficient solutions \citep[e.g., via cognitively plausible search heuristics, or implementing amortization of the reasoning, cf.][]{gershman2014amortized}.
Further, likelihood estimates supplied by the LLMs could be used for evaluations instead of the sampling based approximation here \citep[cf.,][]{hu2023prompt}. This would allow for quantitative model comparison \citep{LeeWagenmakers2013:Bayesian-Cognit,FrankeTsvilodub2024:Bayesian-Statis}. 
Finally, neural modules implementing specific subtasks can be reused across different cognitive models. For instance, the UtterancesProposer module could be reused for sampling plausible utterances in other tasks or contexts. Repeatedly deployed and evaluated modules could over time accumulate into a toolbox of well-tested modules for cognitive modeling. Modularity is cognitive motivated \citep[e.g.,][]{Sperber2001-SPEIDO-2}, and additionally modules for common sub-tasks could be fine-tuned, which might become increasingly important as more agent models and cognitive architectures are deployed \citep[cf.][]{sumers2023cognitive}.

In sum, we take this case study to be an informative starting point outlining some methodological reference points for further investigating the potential of explanatory cognitive models augmented with neural modules within the toolbox of cognitive scientists.

\newpage

\section*{Acknowledgements}
We gratefully acknowledge support by the state of Baden-Württemberg, Germany, through the computing resources provided by bwHPC and
the German Research Foundation (DFG) through grant INST
35/1597-1 FUGG. MF is a member of the Machine Learning
Cluster of Excellence, EXC number 2064/1 – Project number 39072764.

\section*{Impact Statement / Limitations}

Our approach also has various limitations. The model crucially depends on the quality of the LLM-based modules, especially since small errors in early iterations might be accumulated and propagated. For example, we observed that evaluation of literal semantics should be critically assessed, and SOTA LLMs tend towards verbose generations without explicit instructions (see Appendix~\ref{app:im-utt-proposer}, \ref{app:im-sem-eval} for details). Manual inspection of the model outputs revealed that, when approaching the maximal number of iterations, utterances sometimes became more repetitive, but remained true. 
LLM modules also inherit issues of LLMs, such as excessive sensitivity to apparently minor changes to the prompt, uninterpretability, hallucinations and biases \cite{bender2021stochastic, ji2023hallucination, liu2023lost, shi2023large, zhao2023explainability}.
Further, they do not perform equally well across different domains \citep[e.g.,][]{ahn-etal-2024-large}; modeling language cognition with such hybrid models where LLMs are employed ``in-distribution'' might therefore be a particularly natural starting point.
An important goal for future work is also to analyze the performance of the modules backed by different (open-source) LLMs; this case study can be seen as a reference point of performance with one of the best available closed-source models.

\nocite{langley00}

\bibliography{example_paper}

\begin{thebibliography}{50}
\providecommand{\natexlab}[1]{#1}
\providecommand{\url}[1]{\texttt{#1}}
\expandafter\ifx\csname urlstyle\endcsname\relax
  \providecommand{\doi}[1]{doi: #1}\else
  \providecommand{\doi}{doi: \begingroup \urlstyle{rm}\Url}\fi

\bibitem[Ahn et~al.(2024)Ahn, Verma, Lou, Liu, Zhang, and Yin]{ahn-etal-2024-large}
Ahn, J., Verma, R., Lou, R., Liu, D., Zhang, R., and Yin, W.
\newblock Large language models for mathematical reasoning: Progresses and challenges.
\newblock In Falk, N., Papi, S., and Zhang, M. (eds.), \emph{Proceedings of the 18th Conference of the European Chapter of the Association for Computational Linguistics: Student Research Workshop}, pp.\  225--237, St. Julian{'}s, Malta, March 2024. Association for Computational Linguistics.
\newblock URL \url{https://aclanthology.org/2024.eacl-srw.17}.

\bibitem[Andreas \& Klein(2016)Andreas and Klein]{AndreasKlein2016:Reasoning-about}
Andreas, J. and Klein, D.
\newblock Reasoning about pragmatics with neural listeners and speakers.
\newblock In \emph{Proceedings of the 2016 Conference on Empirical Methods in Natural Language Processing}, pp.\  1173--1182, Austin, TX, 2016. Association for Computational Linguistics.

\bibitem[Bender et~al.(2021)Bender, Gebru, McMillan-Major, and Mitchell]{bender2021stochastic}
Bender, E.~M., Gebru, T., McMillan-Major, A., and Mitchell, M.
\newblock On the dangers of stochastic parrots: Can language models be too big?
\newblock In \emph{Proceedings of the 2021 ACM Conference on Fairness, Accountability, and Transparency}, FAccT '21, pp.\  610–623, New York, NY, USA, 2021. Association for Computing Machinery.
\newblock ISBN 9781450383097.
\newblock \doi{10.1145/3442188.3445922}.
\newblock URL \url{https://doi.org/10.1145/3442188.3445922}.

\bibitem[Bommasani et~al.(2021)Bommasani, Hudson, Adeli, Altman, Arora, von Arx, Bernstein, Bohg, Bosselut, Brunskill, et~al.]{bommasani2021opportunities}
Bommasani, R., Hudson, D.~A., Adeli, E., Altman, R., Arora, S., von Arx, S., Bernstein, M.~S., Bohg, J., Bosselut, A., Brunskill, E., et~al.
\newblock On the opportunities and risks of foundation models.
\newblock \emph{arXiv preprint arXiv:2108.07258}, 2021.

\bibitem[Bowman et~al.(2015)Bowman, Angeli, Potts, and Manning]{bowman-etal-2015-large}
Bowman, S.~R., Angeli, G., Potts, C., and Manning, C.~D.
\newblock A large annotated corpus for learning natural language inference.
\newblock In M{\`a}rquez, L., Callison-Burch, C., and Su, J. (eds.), \emph{Proceedings of the 2015 Conference on Empirical Methods in Natural Language Processing}, pp.\  632--642, Lisbon, Portugal, September 2015. Association for Computational Linguistics.
\newblock \doi{10.18653/v1/D15-1075}.
\newblock URL \url{https://aclanthology.org/D15-1075}.

\bibitem[Brown et~al.(2020)Brown, Mann, Ryder, Subbiah, Kaplan, Dhariwal, Neelakantan, Shyam, Sastry, Askell, Agarwal, Herbert-Voss, Krueger, Henighan, Child, Ramesh, Ziegler, Wu, Winter, Hesse, Chen, Sigler, Litwin, Gray, Chess, Clark, Berner, McCandlish, Radford, Sutskever, and Amodei]{NEURIPS2020_1457c0d6}
Brown, T., Mann, B., Ryder, N., Subbiah, M., Kaplan, J.~D., Dhariwal, P., Neelakantan, A., Shyam, P., Sastry, G., Askell, A., Agarwal, S., Herbert-Voss, A., Krueger, G., Henighan, T., Child, R., Ramesh, A., Ziegler, D., Wu, J., Winter, C., Hesse, C., Chen, M., Sigler, E., Litwin, M., Gray, S., Chess, B., Clark, J., Berner, C., McCandlish, S., Radford, A., Sutskever, I., and Amodei, D.
\newblock Language models are few-shot learners.
\newblock In Larochelle, H., Ranzato, M., Hadsell, R., Balcan, M., and Lin, H. (eds.), \emph{Advances in Neural Information Processing Systems}, volume~33, pp.\  1877--1901. Curran Associates, Inc., 2020.

\bibitem[Burgess \& Kim(2018)Burgess and Kim]{burgess20183d}
Burgess, C. and Kim, H.
\newblock 3d shapes dataset, 2018.

\bibitem[Chowdhery et~al.(2022)Chowdhery, Narang, Devlin, Bosma, Mishra, Roberts, Barham, Chung, Sutton, Gehrmann, et~al.]{chowdhery2022palm}
Chowdhery, A., Narang, S., Devlin, J., Bosma, M., Mishra, G., Roberts, A., Barham, P., Chung, H.~W., Sutton, C., Gehrmann, S., et~al.
\newblock Palm: Scaling language modeling with pathways.
\newblock \emph{arXiv preprint arXiv:2204.02311}, 2022.

\bibitem[Cohn-Gordon et~al.(2018)Cohn-Gordon, Goodman, and Potts]{Cohn-GordonGoodman2018:Pragmatically-I}
Cohn-Gordon, R., Goodman, N.~D., and Potts, C.
\newblock Pragmatically informative image captioning with character-level inference.
\newblock In \emph{Proceedings of the 2018 Conference of the {N}orth {A}merican Chapter of the {A}ssociation for {C}omputational {L}inguistics: Human Language Technologies}, pp.\  439--443, Stroudsburg, PA, June 2018. Association for Computational Linguistics.

\bibitem[Creswell et~al.(2022)Creswell, Shanahan, and Higgins]{creswell2022selection}
Creswell, A., Shanahan, M., and Higgins, I.
\newblock Selection-inference: Exploiting large language models for interpretable logical reasoning.
\newblock \emph{arXiv preprint arXiv:2205.09712}, 2022.

\bibitem[Dale \& Reiter(1995)Dale and Reiter]{dale1995computational}
Dale, R. and Reiter, E.
\newblock Computational interpretations of the gricean maxims in the generation of referring expressions.
\newblock \emph{Cognitive science}, 19\penalty0 (2):\penalty0 233--263, 1995.

\bibitem[Ferreira(2019)]{Ferreira2019:A-Mechanistic-F}
Ferreira, V.~S.
\newblock A mechanistic framework for explaining audience design in language production.
\newblock \emph{Annual Review of Psychology}, 70\penalty0 (1):\penalty0 29--51, 2019.

\bibitem[Frank \& Goodman(2012)Frank and Goodman]{frank2012predicting}
Frank, M.~C. and Goodman, N.~D.
\newblock Predicting pragmatic reasoning in language games.
\newblock \emph{Science}, 336\penalty0 (6084):\penalty0 998--998, 2012.

\bibitem[Franke et~al.(2024)Franke, Tsvilodub, and Carcassi]{FrankeTsvilodub2024:Bayesian-Statis}
Franke, M., Tsvilodub, P., and Carcassi, F.
\newblock {Bayesian Statistical Modeling with Predictors from LLMs}, 2024.

\bibitem[Gao et~al.(2022)Gao, Madaan, Zhou, Alon, Liu, Yang, Callan, and Neubig]{gao2022pal}
Gao, L., Madaan, A., Zhou, S., Alon, U., Liu, P., Yang, Y., Callan, J., and Neubig, G.
\newblock Pal: Program-aided language models.
\newblock \emph{arXiv preprint arXiv:2211.10435}, 2022.

\bibitem[Gershman \& Goodman(2014)Gershman and Goodman]{gershman2014amortized}
Gershman, S. and Goodman, N.
\newblock Amortized inference in probabilistic reasoning.
\newblock In \emph{Proceedings of the annual meeting of the cognitive science society}, volume~36, 2014.

\bibitem[Grice(1975)]{grice1975logic}
Grice, H.~P.
\newblock Logic and conversation.
\newblock In \emph{Speech acts}, pp.\  41--58. Brill, 1975.

\bibitem[He-Yueya et~al.(2023)He-Yueya, Poesia, Wang, and Goodman]{he2023solving}
He-Yueya, J., Poesia, G., Wang, R.~E., and Goodman, N.~D.
\newblock Solving math word problems by combining language models with symbolic solvers.
\newblock \emph{arXiv preprint arXiv:2304.09102}, 2023.

\bibitem[Hendricks et~al.(2016)Hendricks, Akata, Rohrbach, Donahue, Schiele, and Darrell]{HendricksAkata2016:Generating-Visu}
Hendricks, L.~A., Akata, Z., Rohrbach, M., Donahue, J., Schiele, B., and Darrell, T.
\newblock Generating visual explanations.
\newblock In Leibe, B., Matas, J., Sebe, N., and Welling, M. (eds.), \emph{Computer Vision -- ECCV 2016}, pp.\  3--19, Cham, 2016. Springer International Publishing.

\bibitem[Hu \& Levy(2023)Hu and Levy]{hu2023prompt}
Hu, J. and Levy, R.
\newblock Prompt-based methods may underestimate large language models' linguistic generalizations.
\newblock \emph{arXiv preprint arXiv:2305.13264}, 2023.

\bibitem[Ji et~al.(2023)Ji, Lee, Frieske, Yu, Su, Xu, Ishii, Bang, Madotto, and Fung]{ji2023hallucination}
Ji, Z., Lee, N., Frieske, R., Yu, T., Su, D., Xu, Y., Ishii, E., Bang, Y.~J., Madotto, A., and Fung, P.
\newblock Survey of hallucination in natural language generation.
\newblock \emph{ACM Comput. Surv.}, 55\penalty0 (12), mar 2023.
\newblock ISSN 0360-0300.
\newblock \doi{10.1145/3571730}.
\newblock URL \url{https://doi.org/10.1145/3571730}.

\bibitem[Kramer \& van Deemter(2012)Kramer and van Deemter]{KramerDeemter2012:Computational-G}
Kramer, E. and van Deemter, K.
\newblock Computational generation of referring expressions: {A} survey.
\newblock \emph{Computational Linguistics}, 38\penalty0 (1):\penalty0 173--218, 2012.

\bibitem[Langley(2000)]{langley00}
Langley, P.
\newblock Crafting papers on machine learning.
\newblock In Langley, P. (ed.), \emph{Proceedings of the 17th International Conference on Machine Learning (ICML 2000)}, pp.\  1207--1216, Stanford, CA, 2000. Morgan Kaufmann.

\bibitem[Lee \& Wagenmakers(2015)Lee and Wagenmakers]{LeeWagenmakers2013:Bayesian-Cognit}
Lee, M.~D. and Wagenmakers, E.-J.
\newblock \emph{Bayesian Cognitive Modeling: {A} Practical Course}.
\newblock Cambridge University Press, Cambridge, MA, 2015.

\bibitem[Lew et~al.(2020)Lew, Tessler, Mansinghka, and Tenenbaum]{lew2020leveraging}
Lew, A.~K., Tessler, M.~H., Mansinghka, V.~K., and Tenenbaum, J.~B.
\newblock Leveraging unstructured statistical knowledge in a probabilistic language of thought.
\newblock In \emph{Proceedings of the annual conference of the cognitive science society}, 2020.

\bibitem[Lewis et~al.(2020)Lewis, Perez, Piktus, Petroni, Karpukhin, Goyal, K{\"u}ttler, Lewis, Yih, Rockt{\"a}schel, et~al.]{lewis2020retrieval}
Lewis, P., Perez, E., Piktus, A., Petroni, F., Karpukhin, V., Goyal, N., K{\"u}ttler, H., Lewis, M., Yih, W.-t., Rockt{\"a}schel, T., et~al.
\newblock Retrieval-augmented generation for knowledge-intensive nlp tasks.
\newblock \emph{Advances in Neural Information Processing Systems}, 33:\penalty0 9459--9474, 2020.

\bibitem[Liu et~al.(2022)Liu, Liu, Lu, Welleck, West, Le~Bras, Choi, and Hajishirzi]{liu-etal-2022-generated}
Liu, J., Liu, A., Lu, X., Welleck, S., West, P., Le~Bras, R., Choi, Y., and Hajishirzi, H.
\newblock Generated knowledge prompting for commonsense reasoning.
\newblock In \emph{Proceedings of the 60th Annual Meeting of the Association for Computational Linguistics (Volume 1: Long Papers)}, pp.\  3154--3169, Dublin, Ireland, 2022. Association for Computational Linguistics.
\newblock \doi{10.18653/v1/2022.acl-long.225}.
\newblock URL \url{https://aclanthology.org/2022.acl-long.225}.

\bibitem[Liu et~al.(2023)Liu, Lin, Hewitt, Paranjape, Bevilacqua, Petroni, and Liang]{liu2023lost}
Liu, N.~F., Lin, K., Hewitt, J., Paranjape, A., Bevilacqua, M., Petroni, F., and Liang, P.
\newblock Lost in the middle: How language models use long contexts, 2023.

\bibitem[{Mao} et~al.(2016){Mao}, {Huang}, {Toshev}, {Camburu}, {Yuille}, and {Murphy}]{MaoHuang2016:Generation-and-}
{Mao}, J., {Huang}, J., {Toshev}, A., {Camburu}, O., {Yuille}, A., and {Murphy}, K.
\newblock Generation and comprehension of unambiguous object descriptions.
\newblock In \emph{2016 IEEE Conference on Computer Vision and Pattern Recognition (CVPR)}, pp.\  11--20, 2016.

\bibitem[Monroe \& Potts(2015)Monroe and Potts]{MonroePotts2015:Learning-in-the}
Monroe, W. and Potts, C.
\newblock Learning in the {R}ational {S}peech {A}cts model.
\newblock In \emph{Proceedings of 20th {A}msterdam {C}olloquium}, Amsterdam, December 2015. ILLC.

\bibitem[Moskal et~al.(2024)Moskal, Musuvathi, and {K\i c\i man}]{Moskal2024}
Moskal, M., Musuvathi, M., and {K\i c\i man}, E.
\newblock {AI Controller Interface}.
\newblock \url{https://github.com/microsoft/aici/}, 2024.

\bibitem[Newell \& Simon(1972)Newell and Simon]{NewellSimon1972a}
Newell, A. and Simon, H.~A.
\newblock \emph{Human problem solving}.
\newblock Prentice-Hall, Englewood Cliffs, NJ, 1972.

\bibitem[Nie et~al.(2020)Nie, Cohn-Gordon, and Potts]{NieCohn-Gordon2020:Pragmatic-Issue}
Nie, A., Cohn-Gordon, R., and Potts, C.
\newblock Pragmatic issue-sensitive image captioning.
\newblock In \emph{Findings of the Association for Computational Linguistics: EMNLP 2020}, pp.\  1924--1938, Online, November 2020. Association for Computational Linguistics.

\bibitem[Ohmer et~al.(2021)Ohmer, Franke, and K{\"o}nig]{OhmerFranke2021:Mutual-Exclusive-CogSci}
Ohmer, X., Franke, M., and K{\"o}nig, P.
\newblock Mutual exclusivity in pragmatic agents.
\newblock \emph{Cognitive Science}, 46\penalty0 (1):\penalty0 e13069, 2021.

\bibitem[Paranjape et~al.(2023)Paranjape, Lundberg, Singh, Hajishirzi, Zettlemoyer, and Ribeiro]{paranjape2023art}
Paranjape, B., Lundberg, S., Singh, S., Hajishirzi, H., Zettlemoyer, L., and Ribeiro, M.~T.
\newblock Art: Automatic multi-step reasoning and tool-use for large language models.
\newblock \emph{arXiv preprint arXiv:2303.09014}, 2023.

\bibitem[Piriyakulkij et~al.(2023)Piriyakulkij, Kuleshov, and Ellis]{piriyakulkij2023asking}
Piriyakulkij, T., Kuleshov, V., and Ellis, K.
\newblock Asking clarifying questions using language models and probabilistic reasoning.
\newblock In \emph{NeurIPS 2023 Foundation Models for Decision Making Workshop}, 2023.
\newblock URL \url{https://openreview.net/forum?id=2SjoG6lVz3}.

\bibitem[Poesia et~al.(2023)Poesia, Gandhi, Zelikman, and Goodman]{poesia2023certified}
Poesia, G., Gandhi, K., Zelikman, E., and Goodman, N.~D.
\newblock Certified deductive reasoning with language models, 2023.

\bibitem[Shi et~al.(2023)Shi, Chen, Misra, Scales, Dohan, Chi, Schärli, and Zhou]{shi2023large}
Shi, F., Chen, X., Misra, K., Scales, N., Dohan, D., Chi, E., Schärli, N., and Zhou, D.
\newblock Large language models can be easily distracted by irrelevant context, 2023.

\bibitem[Sperber(2001)]{Sperber2001-SPEIDO-2}
Sperber, D.
\newblock In defense of massive modularity.
\newblock In Dupoux, E. (ed.), \emph{Language, Brain and Cognitive Development: Essays in Honor of Jacques Mehler}. MIT Press, 2001.

\bibitem[Sumers et~al.(2023)Sumers, Yao, Narasimhan, and Griffiths]{sumers2023cognitive}
Sumers, T., Yao, S., Narasimhan, K., and Griffiths, T.~L.
\newblock Cognitive architectures for language agents.
\newblock \emph{arXiv preprint arXiv:2309.02427}, 2023.

\bibitem[Touvron et~al.(2023)Touvron, Lavril, Izacard, Martinet, Lachaux, Lacroix, Rozi{\`e}re, Goyal, Hambro, Azhar, et~al.]{touvron2023llama}
Touvron, H., Lavril, T., Izacard, G., Martinet, X., Lachaux, M.-A., Lacroix, T., Rozi{\`e}re, B., Goyal, N., Hambro, E., Azhar, F., et~al.
\newblock Llama: Open and efficient foundation language models.
\newblock \emph{arXiv preprint arXiv:2302.13971}, 2023.

\bibitem[Tsvilodub \& Franke(2023)Tsvilodub and Franke]{tsvilodub-franke-2023-evaluating}
Tsvilodub, P. and Franke, M.
\newblock Evaluating pragmatic abilities of image captioners on {A}3{DS}.
\newblock In Rogers, A., Boyd-Graber, J., and Okazaki, N. (eds.), \emph{Proceedings of the 61st Annual Meeting of the Association for Computational Linguistics (Volume 2: Short Papers)}, pp.\  1277--1285, Toronto, Canada, 2023. Association for Computational Linguistics.
\newblock \doi{10.18653/v1/2023.acl-short.110}.
\newblock URL \url{https://aclanthology.org/2023.acl-short.110}.

\bibitem[{v}an Rooij et~al.(2023){v}an Rooij, Guest, Adolfi, de~Haan, Kolokolova, and Rich]{van2023reclaiming}
{v}an Rooij, I., Guest, O., Adolfi, F.~G., de~Haan, R., Kolokolova, A., and Rich, P.
\newblock Reclaiming {AI} as a theoretical tool for cognitive science.
\newblock 2023.

\bibitem[Vedantam et~al.(2017)Vedantam, Bengio, Murphy, Parikh, and Chechik]{VedantamBengio2017:Context-Aware-C}
Vedantam, R., Bengio, S., Murphy, K., Parikh, D., and Chechik, G.
\newblock Context-aware captions from context-agnostic supervision.
\newblock In \emph{Proceedings of the IEEE Conference on Computer Vision and Pattern Recognition (CVPR)}, pp.\  251--260, 2017.

\bibitem[Wang et~al.(2019)Wang, Pruksachatkun, Nangia, Singh, Michael, Hill, Levy, and Bowman]{wang2019superglue}
Wang, A., Pruksachatkun, Y., Nangia, N., Singh, A., Michael, J., Hill, F., Levy, O., and Bowman, S.
\newblock Superglue: A stickier benchmark for general-purpose language understanding systems.
\newblock \emph{Advances in neural information processing systems}, 32, 2019.

\bibitem[Wei et~al.(2022)Wei, Wang, Schuurmans, Bosma, Chi, Le, and Zhou]{wei2022chain}
Wei, J., Wang, X., Schuurmans, D., Bosma, M., Chi, E., Le, Q., and Zhou, D.
\newblock Chain of thought prompting elicits reasoning in large language models.
\newblock \emph{arXiv preprint arXiv:2201.11903}, 2022.

\bibitem[Wong et~al.(2023)Wong, Grand, Lew, Goodman, Mansinghka, Andreas, and Tenenbaum]{wong2023word}
Wong, L., Grand, G., Lew, A.~K., Goodman, N.~D., Mansinghka, V.~K., Andreas, J., and Tenenbaum, J.~B.
\newblock From word models to world models: Translating from natural language to the probabilistic language of thought, 2023.

\bibitem[Yao et~al.(2023)Yao, Yu, Zhao, Shafran, Griffiths, Cao, and Narasimhan]{yao2023tree}
Yao, S., Yu, D., Zhao, J., Shafran, I., Griffiths, T.~L., Cao, Y., and Narasimhan, K.
\newblock Tree of thoughts: Deliberate problem solving with large language models.
\newblock \emph{arXiv preprint arXiv:2305.10601}, 2023.

\bibitem[Zarrie{\ss} \& Schlangen(2019)Zarrie{\ss} and Schlangen]{ZarriessSchlangen2019:Know-What-You-D}
Zarrie{\ss}, S. and Schlangen, D.
\newblock Know what you don{'}t know: Modeling a pragmatic speaker that refers to objects of unknown categories.
\newblock In \emph{Proceedings of the 57th Annual Meeting of the Association for Computational Linguistics}, pp.\  654--659. Association for Computational Linguistics, 2019.

\bibitem[Zhao et~al.(2023)Zhao, Chen, Yang, Liu, Deng, Cai, Wang, Yin, and Du]{zhao2023explainability}
Zhao, H., Chen, H., Yang, F., Liu, N., Deng, H., Cai, H., Wang, S., Yin, D., and Du, M.
\newblock Explainability for large language models: A survey.
\newblock \emph{ACM Transactions on Intelligent Systems and Technology}, 2023.

\end{thebibliography}
\bibliographystyle{icml2024}

\newpage
\appendix

\section{Implementation and Evaluation Details}
\label{app:section}

\subsection{Iterative Model Details}

\begin{algorithm}[t]
\caption{Iterative model. Components in \llm{brown} are implemented as LLM modules in this case study, components in \symbolic{green} are symbolic. }\label{alg:iterative-model}
\begin{algorithmic}
    \STATE \texttt{Generate($s^{*}$, $D$, $n$)}:
    \STATE $\text{partialUtt} \gets []$
    \WHILE{True}
        \FOR{$u' \in \text{partialUtt}$}
        \STATE $U \gets \llm{\text{UtterancesProposer}}(s^{*}, n, u')$
        \ENDFOR
        \STATE $C_{\text{new}} \gets []$
        \STATE $T \gets []$
        \FOR{$u \in U$}
            \STATE \textbf{append} $ T_u = \llm{\text{SemanticEvaluator}}(\{s^{*}\} \cup D, u)$ \textbf{to} $T$
            \STATE \textbf{append} $\symbolic{\text{ConstrastivitySelector}}(T_u)$ \textbf{to} $C_{\text{new}}$
        \ENDFOR
        \STATE $\text{indices} \gets i \mid c_i \in \max (C_{\text{new}})$
        \STATE $U^* \gets U[\text{indices}]$
        \STATE $T^* \gets T[\text{indices}]$
        \IF{$\max (C_{\text{new}}) = 1$} 
            \STATE $u^* \gets \symbolic{\text{InfoMaxSelector}}(T^*, U^*)$
            \STATE \textbf{return} $u^*$
        \ENDIF
        \FOR{$u \in U^*$}
            \STATE partialUtt $\gets U^*$
            \STATE $U \gets []$ 
        \ENDFOR
    \ENDWHILE
\end{algorithmic}
\label{alg:utteranceProductionLoopFlow}
\end{algorithm}

More technically, IM (Algorithm~\ref{alg:iterative-model}) takes as input a list of full state descriptions, one of which is the target state and the remaining ones are distractors. The target is always passed as the first state. First, the (LLM-based) UtterancesProposer generates candidate utterances that describe a single detail of the target state based on the target state description. Second, the SemanticEvaluator determines the (literal) truth value of all candidate utterances for each state (target and distractors). Third, based on the semantic evaluation in the previous step, the ContrastivitySelector evaluates the contrastivity of the generated utterances, and determines whether any utterance is fully contrastive (i.e., only true of the target). For each utterance $u_i$, this is computed as $C_i = 1 - \frac{\# \text{distractors for which } [\![u_i]\!](d) = 1}{\# \text{distractors}}$. A fully contrastive utterance has $C_i = 1$.
If utterances with $C_i=1$ are found, one utterance is greedily chosen among the contrastive utterances by the InfoMaxSelector and returned. 

Otherwise, a set of the so far most contrastive utterances is constructed, and the passed to the UtterancesProposer module again. This module produces new alternatives for each utterance, each of which includes one more detail from the full description. The loop repeats from the semantic evaluation on until a fully contrastive utterance is produced or the maximal iteration steps have been reached, in which case the InfoMaxSelector greedily selects the most contrastive among the utterances produced in the most recent iteration.

\subsubsection{UtterancesProposer}
\label{app:im-utt-proposer}
On the first iteration of the model, there are no previous utterances yet. We prompt the LLM to generate initial utterances which should only mention a single feature of the target. This prompt can be considered as simulating a production cost pressure, which is present for humans and usually included in cognitive models, but absent in LLMs.
Based on our explorations, we observe that for \texttt{gpt-3.5-turbo} the prompt needs to explicitly instruct the LLM to NOT generate exhaustive utterances. 
We used the following prompt (A) for the first iteration:
\begin{prompt}
    \noindent You will be given a target description. Please produce \{num\_samples\} sentence(s) that only mention one detail from the target description. The produced sentences should include exclusively content mentioned in the target. Please provide the sentences in a bullet list format.  \\

    Target: \{target state\}\\
    Sentences:
\end{prompt}

\texttt{num\_samples} was set to either four or eight in our simulation.

For further iterations of the model, the prompt (B) was adapted to also include previously generated utterances:
\begin{prompt}
    \noindent Your task is to produce some sentences. Each sentence should repeat the information in "\{partial\_description\}", but add one more detail taken from "\{full\_description\}" Do not make up any new detail. \\

    Please produce \{num\_samples\} sentence(s) in a bullet list format. Be very concise!\\

    Sentences:
\end{prompt}

\paragraph{Evaluation} Samples from all of the reference game were manually inspected by the authors and no grammatical errors were observed. It was observed that in some of the samples more than one feature of the target was added in one iteration. For instance, the object type (i.e,. the shape) was mentioned in addition to some other new feature, akin to a basic-level bias.  

\subsubsection{SemanticEvaluator}
\label{app:im-sem-eval}
The purpose of the SemanticEvaluator is to check whether the candidate utterances can be used to refer to each of the states (target and distractors). The task is cast as checking the utterance semantics and checking if the utterance is literally true of the state. Semantic evaluation is applied to each utterance-state combination.

Based on a state description, an utterance and a prompt, the LLM is instructed to answer the question contained in the prompt with `yes'/`no' and output a chain of thought. The generated response is processed with a regular expression to extract `yes'/`no' and convert these to 1, 0, respectively. We note that since the time of implementation, new methods for retrieving LLM generations that fit a specific format and contain specific values have been developed \citep[e.g.,][]{Moskal2024}; such tools could be of great use for future implementations of such modules. 

To exploit the model's knowledge of intuitive language use, the prompt chosen to determine semantic compatibility can be seen as implementing an \emph{intuitive entailment} task. The prompt was:

\begin{prompt}
Consider the following sentence: \{state\}

Does the following statement provide exclusively information also contained in the sentence above: \{utterance\} 

Explain your answer step by step. 

Importantly, the last line of your answer should exclusively contain "yes" or "no", and nothing else. 

Here the the structure of the answer:

"""

[step-by-step explanation,

possibly over multiple lines]

[empty line]

[yes/no]

"""

Your answer: 
\end{prompt}
During development, variations of this prompt were tested. For instance, we added a one-shot chain of thought prompt exemplifying the reasoning (no effect observed); various terms used to refer to the state and the previous utterance were used (``statement'', ``sentence'', ``fact''). Alternative more natural-sounding prompt was tested:
\begin{prompt}
Suppose you already know the following facts: \{state\}

Do you learn anything new from the following statement: \{utterance\}?

Explain your answer step by step.
\end{prompt}

The final prompt presented above led to best results during testing and was, therefore, used throughout simulations for both the IM and the SP model.

\paragraph{Evaluation}
Since the semantic evaluation crucially carries the performance of the entire model, we analysed the performance of this module in isolation. 
Specifically, we evaluated the SemanticEvaluator on two groups of tests. First, we used tasks from the SuperGLUE and SNLI benchmarks \cite{bowman-etal-2015-large, wang2019superglue}. One set of tests is based on five samples from each of the ``axb'', ``axg'', ``copa'', ``rte'' tasks within the SuperGLUE benchmark, and five entailment and contradiction datapoints each from the SNLI benchmark. Pairs of states and utterances where constructed, and the ground truth semantic value was derived from the dataset values. For SNLI, the sentence 1 was used as the state and sentence 2 as the utterance. These tests mostly contained naturalistic sentences and strongly focused on testing NLI. There were 39 test pairs.
Second, we used tests containing example sentences closely matching A3DS in phrasing and content, checking for synonym and modification understanding, as well as some further semantic tests with examples of quantifiers. These are intended to broaden the set of tests and include tests matching the reference game setting more closely. There were 12 test sentences. 

All evaluations were conducted with manual evaluation.
The prompt was optimized based on performance on these tests. The accuracy of the final SemanticEvaluator with the prompt reported above was 0.82.
Inspecting cases where the module failed throughout development, we identified some systematic ways in which this prompt fails. 
For instance, if the chain of thought of the model seems to indicate that the LLM answers the question ``Is there any new information \textit{present} in the utterance?'', it leads the model to incorrectly answer with ``no'', although the reasoning suggests ``yes''. 


\subsubsection{ContrastivitySelector}
This module is a rule-based module which takes a matrix of truth values $T$ computed by the semantic evaluator (i.e., a matrix of shape state$\times$utterance), and checks the proportion of distractors of which each available utterance is \emph{false}. That is, it computes the sum $S$ of the truth values over the states and computes $C = 1 - softmax(S)$, resulting in a list of contrastivity values for the utterances.
If there is at least one utterance which is fully contrastive (i.e., $max(C) = 1$), the loop is terminated and this utterance is returned. If no utterance is fully contrastive, the utterances with the highest contrastivity are selected and passed to the extended UtterancesProposer.

\subsubsection{InfoMaxSelector}
\label{app:rsa-reasoner}
We implement a simple informativity maximization (InfoMax) utterance selector which returns the utterance with the highest contrastivity value, which is derived from the semantic truth values.

Specifically, the selector takes as input the result of the ContrastivityEvaluator and selects the optimal contrastive utterance $u$:
$u^* = argmax (C)$

We note that this module could be extended to return a distribution over utterances, akin to the pragmatic speaker $S_1$ in RSA models \cite{frank2012predicting}:

\begin{align}
P_{S_1}(u \mid s) 
&\propto \exp(\alpha \; (\log \; L_0(s \mid u) - \text{cost})) \\
P_{L_0}(s\mid u) 
&\propto [\![u]\!](s)
\end{align}

\subsection{Ablated Single-Pass Model Details}

As for the IM, the input to the model is a list of state descriptions, including the target state and one or more distractors. The model proceeds in three steps (Algorithm~\ref{alg:ablated-model}). First, an UtterancesProposer module generates ten candidate utterances for the target state based on the target state description. Second, the SemanticEvaluator module determines the truth value of each candidate utterance for each state (target and distractors). Lastly, the InfoMaxSelector module selects the most contrastive  utterance. The functionality of the SemanticEvalutor and the InfoMaxSelector are identical to the IM.

\subsubsection{UtterancesProposer}
The prompt used for generating the utterance proposals did not include the restriction to mentioning a single feature, i.e., it did not include the production cost approximation.

\begin{prompt}
    You will be given a target description. Please produce \{num\_samples\} sentence(s) based on the target that leave out some part of the description but are still well-formed. The reduced sentences should include exclusively content already mentioned in Target. Please provide the sentences in a bullet list format.
\end{prompt}

\texttt{\{num\_samples\}} was set to ten across simulations.


\subsection{LLM Baseline}
\label{app:baseline}

We conducted the same simulations with a one-shot CoT LLM baseline, where the LLM was prompted with all context information and was directly prompted to solve the task of generating a contrastive utterance. The following prompt was used:

\begin{prompt}
\noindent You will be given a target state and one or more distractors. \\
Your task is to describe the target state in natural language in a way that distinguishes it from the distractors. \\
Try to be as concise as possible. You do not need to list all the features of the target state. \\
Please think step by step, motivating why you decide to mention some features.\\

Here is an example of a good answer.\\

Target state:\\
- The floor is purple, the wall is green, the red small block is in the left corner.\\

Distractors:\\
- The floor is red, the wall is green, the red small block is in the middle.\\

Your answer:\\
One difference between the target and the distractor is the color. This difference is enough to distinguish between them.
Utterance: "The target state has a purple floor".\\
Now the real input.
\end{prompt}

\begin{algorithm}[h]
\caption{Ablated (i.e., single-pass) model.}\label{alg:ablated-model} 
\begin{algorithmic}
\STATE \texttt{Generate}{($s^{*}$, $D$, $n$):} 
    \STATE $U_{*} \gets \llm{\text{UtterancesProposer}}(s^{*}, n)$ 
    \STATE $T \gets \llm{\text{SemanticEvaluator}}( \{s^{*}\} \cup D, U_{*})$
    \STATE $u^* \gets \symbolic{\text{InfoMaxSelector}}(T, U_{*})$
    \STATE \textbf{return} $u^*$
\end{algorithmic}
\end{algorithm}

\end{document}